\documentclass{article}

\usepackage[preprint]{nips_2018}
\usepackage[numbers]{natbib}
\usepackage{placeins}
\usepackage[dvipsnames]{xcolor}

\usepackage[utf8]{inputenc} 
\usepackage[T1]{fontenc}    
\usepackage{hyperref}       
\usepackage{url}            
\usepackage{booktabs}       
\usepackage{amsfonts}       
\usepackage{nicefrac}       
\usepackage{microtype}      
\usepackage{graphicx}
\usepackage{amsmath}

\title{Technical Note on Transcription Factor Motif Discovery from Importance Scores (TF-MoDISco) version 0.5.6.5}

\author{
  Avanti Shrikumar\\
  Stanford University\\
  \texttt{avanti@stanford.edu} \\
  \And
  Katherine Tian \\
  The Harker School\\
  \texttt{19katherinet@students.harker.org} \\
  \And
  Žiga Avsec\\
  Technical University of Munich\\
  \texttt{avsec@in.tum.de}\\
  \And
  Anna Shcherbina \\
  Stanford University\\
  \texttt{annashch@stanford.edu} \\
  \And
  Abhimanyu Banerjee\\
  Stanford University\\
  \texttt{manyu@stanford.edu}\\
  \And
  Mahfuza Sharmin\\
  Stanford University\\
  \texttt{msharmin@stanford.edu }\\
  \And
  Surag Nair\\
  Stanford University\\
  \texttt{surag@stanford.edu}\\
  \And
  Anshul Kundaje\\
  Stanford University\\
  \texttt{akundaje@stanford.edu} \\
}

\begin{document}

\maketitle

\begin{abstract}
Deep Learning has recently gained popularity in genomic sequence modeling tasks \cite{alipanahi2015predicting, kelley2016basset, zhou2015predicting}, but interpretation of these models remains challenging. Of particular interest is understanding the recurring patterns learned by the model. Existing motif discovery methods for neural networks visualize individual convolutional filters \cite{lanchantin2016deep, alipanahi2015predicting, kelley2016basset}, but these methods do not account for the fact that deep neural networks learn distributed representations where multiple neurons cooperate to describe a single pattern. Thus, the patterns recognized by individual filters are often found to be partially redundant with each other, hampering interpretability and making it difficult to obtain a non-redundant set of predictive motifs learned by the neural network. In this technical note, we describe TF-MoDISco (Transcription Factor Motif Discovery from Importance Scores), a novel algorithm that leverages per-base importance scores to simultaneously incorporate information from all neurons in the network and generate high-quality, consolidated, non-redundant motifs. This technical note focuses on version 0.5.6.5. The implementation is available at (\url{https://github.com/kundajelab/tfmodisco/tree/v0.5.6.5}).
\end{abstract}

\section{Introduction}

Convolutional neural networks have been used in recent years to successfully learn regulatory patterns in genomic DNA \cite{alipanahi2015predicting,kelley2016basset,lanchantin2016deep}. Combinations of Transcription Factors (TFs) bind combinations of motifs in DNA sequence at non-coding regulatory elements to control gene expression. While the core sequence motifs of a subset of TFs are relatively well-known, the role of flanking nucleotides that influence \textit{in vivo} TF binding, as well as the combinatorial interactions with other TFs, remain largely uncharted. Deep learning models are appealing for this problem because of their ability to learn complex, hierarchical, predictive patterns directly from raw DNA sequence, thus removing the need to explicitly featurize the data (such as featurization using a database of known motifs). Convolutional Neural Networks (CNNs) in particular contain several hierarchical layers of pattern-matching units referred to as convolutional filters that are well suited to learning from DNA sequence. In these models, each convolutional filter in the first layer learns a sequence pattern that is analogous to a position weight matrix (PWM). The filter is scanned (convolved) with the sequence to produce a score for the strength of the match at each position. Later convolutional layers operate on the scores from all filters in the previous layer, allowing the network to learn complex higher-order patterns.

A barrier to the adoption of deep learning models for genomic applications is the difficulty in interpreting the models. While several methods such as \citet{lanchantin2016deep,shrikumar2017learning} and \citet{sundararajan2017axiomatic} have been developed to assign importance scores to each base of an input sequence, methods for learning re-occurring patterns do not leverage importance scores and are largely limited to variations of visualizing the learned representations of individual CNN filters \cite{lanchantin2016deep,kelley2016basset,alipanahi2015predicting}. In practice, this poses an issue because CNNs learn highly distributed representations, meaning that the patterns found by individual convolutional neurons may not be very informative.

In this technical note, we detail the methods behind version 0.5.6.5 of TF-MoDISco. TF-MoDISco is a method for identification of high-quality, consolidated motifs using deep learning. The critical insight of TF-MoDISco is that the importance scores on the inputs are computed using information from all the neurons in the network; thus, by clustering segments of high importance in the inputs, it is possible to identify consolidated motifs learned by the deep learning model. The implementation is available at \url{https://github.com/kundajelab/tfmodisco}, and the release history is available at \url{https://github.com/kundajelab/tfmodisco/releases}. The main difference relative to version 0.4.2.2 (the version described in the previous technical note) is a modification of how the null distribution is handled in \textbf{Section ~\ref{sec:seqletidentification}}.

Examples applications of TF-MoDISco can be found in \citet{gkmexplain} and \citet{bpnet}. This technical note focuses only on providing a description of the methods.

\begin{figure}[!h]
  \begin{center}
  \includegraphics[width=0.95\linewidth]{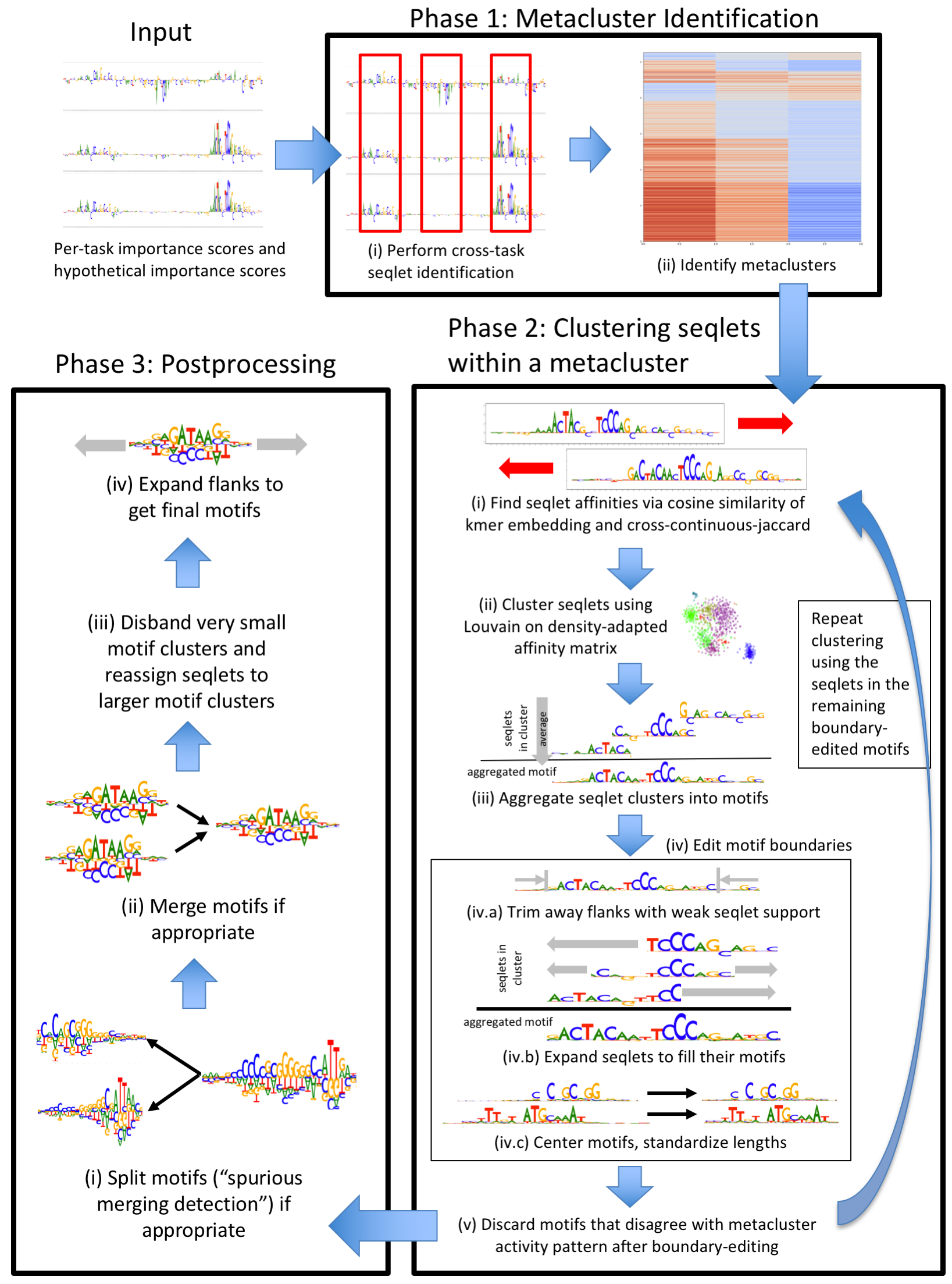}
  \caption[Summary of TF-MoDISco]{\textbf{Summary of TF-MoDISco}}
  \label{fig:methodsumary}
  \end{center}
\end{figure}

\section{Input to TF-MoDISco}
\label{sec:impandhypimp}

TF-MoDISco takes as its input per-base importance scores for every prediction task. These importance scores can be derived through a variety of methods including the ones described in \cite{shrikumar2017learning}. A positive importance value for a particular task indicates that the base influenced the output of the network towards a positive prediction for the task, and a negative importance indicates that the base influences the output of the network towards a negative prediction. Scores that are near zero indicate that the particular base is unimportant for the task in question.

\begin{figure}[!h]
  \begin{center}
  \includegraphics[width=\linewidth]{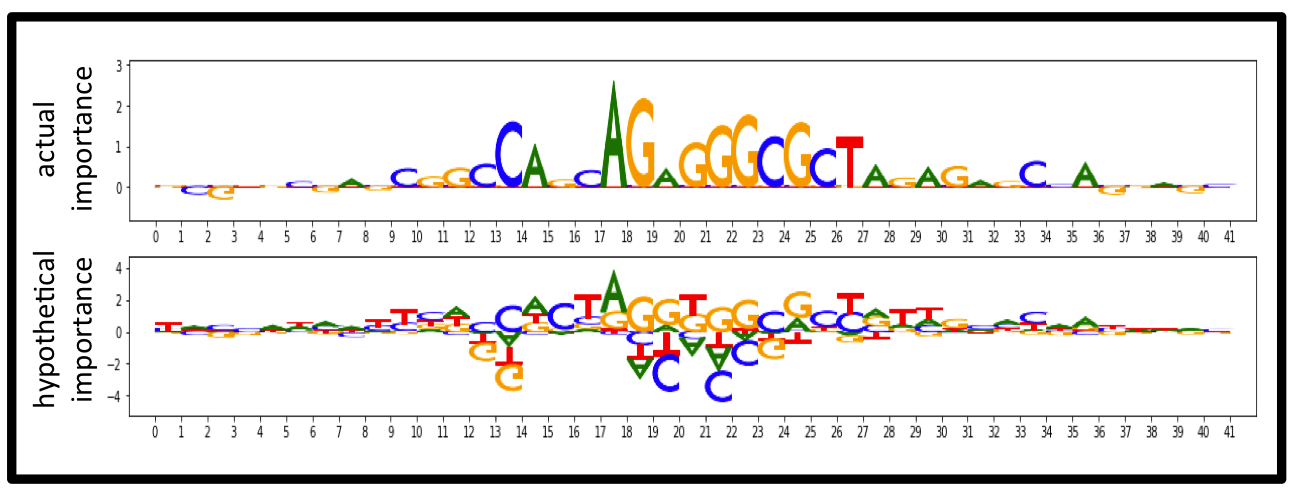}
  \caption[Actual and hypothetical importance scores for CTCF task on an example sequence containing the CTCF motif]{\textbf{Actual and hypothetical importance scores for CTCF task on an example sequence containing the CTCF motif}. The hypothetical importance reveals the impact of other bases not present in the original sequence.}
  \label{fig:hypotheticalimportance}
  \end{center}
\end{figure}

We found that TF-MoDISco results were better if, in addition to using importance scores on the input sequence, we incorporated information about \emph{hypothetical} importance if other unobserved bases were present. A hypothetical importance score answers the question ``if I were to mutate the sequence at a particular position to a different letter, what would the importance on the newly-introduced letter look like?''. As a specific example, consider a basic importance-scoring method such as gradient $\times$ input. When a sequence is one-hot encoded (i.e. `input' can be either 1 or 0), the value of gradient $\times$ input would be zero on all bases that are absent from the sequence and equal to the gradient on the bases that are present (here, `base' refers to a specific letter at a specific position; at every position, only one of ACGT can be present). If a single position were mutated to a different base, one might anticipate that the value of gradient $\times$ input at the newly-introduced base would be close to the current value of the gradient (assuming that the gradient doesn't change dramatically as a result of the mutation). Thus, the gradients give a readout of what the contributions to the network would be if different bases were present at particular positions; if (gradient $\times$ input) is used as the importance scoring method, then the gradients alone would be a good choice for the ``hypothetical importance''. In practice, the ``hypothetical importance'' behaves like an autocomplete of the sequence, giving insight into what patterns the network was looking for at a given region (Fig. ~\ref{fig:hypotheticalimportance}).

What is a good choice of hypothetical importance when DeepLIFT scores are used? DeepLIFT defines quantities known as the multipliers $m_{\Delta x \Delta t}$ such that:

\begin{equation*}
m_{\Delta x \Delta t} = \frac{C_{\Delta x \Delta t}}{\Delta x}
\end{equation*}

where $x$ is the input, $t$ is the output that we are computing contributions to, $\Delta x$ is the change in $x$ (formally the `difference-from-reference') and $C_{\Delta x \Delta t}$ is the contribution of change in $x$ to the change in $t$. If $x^h$ denotes a hypothetical value of the input, then we can approximate $C_{\Delta x^h \Delta t}$ as:

\begin{equation}
C_{\Delta x^h \Delta t} = \Delta x^h m_{\Delta {x^h} \Delta t}
\end{equation}

In other words, we use the multipliers on the current input, but substitute the difference-from-reference of the hypothetical value of the input. Note that if the reference is set to 0 and $m_{\Delta x \Delta t}$ is taken to be the gradient of $t$ w.r.t. $x$, then the formula reduces to the case of gradient $\times$ input described in the previous paragraph. When the reference is not 0 and the input is subject to a one-hot encoding constraint, as is the case with DNA sequence input, care must be taken to project the contributions from the difference-from-reference of all the bases at a position onto the base that is actually present. A jupyter notebook with code illustrating how to obtain importance scores and hypothethetical scores using DeepLIFT on genomic sequences is avaiable at \url{https://github.com/kundajelab/tfmodisco/blob/v0.5.6.5/examples/simulated_TAL_GATA_deeplearning/Generate\%20Importance\%20Scores.ipynb}.

In the case of importance scores derived from a support vector machine, hypothetical importance scores could be obtained as described in \citet{gkmexplain}. Note that when using importance scores from GkmExplain with TF-MoDISco, it is advisable to normalize the scores such that the hypothetical importance at each position sums to 0 across the ACGT bases. The normalization is illustrated in \href{https://github.com/kundajelab/tfmodisco/blob/ffdc70bf08852a18f41ba684a58f54cbd5f8c1d0/examples/H1ESC_Nanog_gkmsvm/TF\%20MoDISco\%20Nanog.ipynb}{\color{blue} this GkmExplain TF-MoDIsco notebook}.

In future versions of TF-MoDISco, we plan to expand the set of allowed input tracks to TF-MoDISco to allow data such as the activations or importance scores on some intermediate convolutional layer, methylation/accessibility information, etc.

\FloatBarrier

\section{TF-MoDISco, Phase 1: Metaclusters}

 The goal of the first phase is to identify segments of the input, termed ``seqlets'', that have substantial contribution to one or more of the output tasks, and to then cluster these seqlets into `metaclusters' such that each have a distinct pattern of contribution to the various tasks. Note that running TF-MoDISco in the multi-task setting is optional; if it produces poor results, we recommend running it on scores from one task at a time.

\subsection{Seqlet identification}
\label{sec:seqletidentification}

After importance scores are obtained, portions of the input with substantial contribution to at least one of the tasks are identified. We refer to these regions as ``seqlets''. A list of seqlet locations is obtained for each individual task and then unified across tasks. Seqlet identification for each individual task proceeds as follows:

\begin{itemize}
    \item For each task, the importance scores are summed in sliding windows of core size \texttt{sliding\_window\_size} with a stride size of 1.
    \begin{itemize}
    	\item If a null distribution of scores is provided (e.g. by computing importance scores on dinucleotide-shuffled sequences), then the sliding window importance scores are also computed on the null.
	\item Otherwise, if a null distribution is not provided, then a Laplace distribution is used the characterize the noise. The Laplace distribution is fit in a way that assumes the low-magnitude scores are mostly noise. First, the mode of the scores is identified, and the scores are re-centered by subtracting the mode. Then, a one-tailed Laplace distribution is fit to each of the 5th, 10th, 15th...85th, 90th \& 95th percentiles of positive window scores. The Laplace distribution with the fastest decrease (i.e. the highest value of lambda) is selected as representative of the null distribution for positive windows. The process is repeated using the negative scores to obtain a null distribution for the negatives windows. We use different Laplace distributions for positive and negative windows in case the positively-valued and negatively-valued noise distributions are asymmetric. Once the Laplace distributions are fit, they are sampled from in order to simulate the result of a user-provided null. 
    \end{itemize}
    \item Based on the null distribution (either determined empirically or by sampling using a Laplace curve), a FDR is calculated at each window score threshold. This is done by estimating the proportion of windows originating from the null distribution at each threshold using isotonic regression. Specifically, an isotonic regression curve is fit separately on positive and negative window scores to distinguish between windows from the null score distribution (labeled 0) and windows from the actual score distribution (labeled 1). The windows from the null score distribution are weighted such that the total weight on all windows from the null score distribution is equal to the total weight on all windows from the original distribution (this corresponds to a prior of 50\% from the null distribution, though the specific choice of prior is less important given that the FDR cutoffs are adjustable by the user). A separate isotonic regression curve is fit for positive windows and negative windows. The smallest-magnitude point on each isotonic regression curve that gives the desired proportion from the null distribution is then used to obtain the positive and negative window thresholds.
    \item If the initial fraction of windows passing the positive and negative thresholds is less than \texttt{min\_passing\_windows\_frac} (set to 0.03 by default) or greater than \texttt{max\_passing\_windows\_frac} (set to 0.2 by default), then the thresholds are readjusted as follows: the windows are sorted in descending order of the absolute value of their importance, and a threshold on the absolute value is selected such that the number of passing windows (relative to the total number of windows) is raised/lowered by the minimum amount needed to meet the bounds specified by \texttt{min\_passing\_windows\_frac} and \texttt{max\_passing\_windows\_frac}. When the threshold is determined in this way (i.e. using absolute values), the threshold for negative windows is equal to -1 multiplied by the threshold for positive windows. If the user does not wish to rely on absolute values and instead wishes to treat the distributions for positive and negative scores separately, they may set the \texttt{separate\_pos\_neg\_thresholds} option to True (it is False by default). If set to True, the fraction of passing positive windows relative to the total number of positive windows would meet the desired target, as would the fraction of passing negative windows relative to the total number of negative windows, but the value of the negative threshold wouldn't be equal to -1 multiplied by the value of the positive threshold unless the distributions were symmetric.
    \item After the final thresholds are determined, all windows that do not pass the threshold are filtered out.
    \item The following process is then repeated for each sequence:
    \begin{itemize}
        \item Identify the window containing a total importance score of the highest magnitude
        \item Expand the window on either side by flank size \texttt{flank\_size}. These coordinates will be used to make a seqlet.
        \item Filter out all windows that would overlap this by over 50\% (after flank expansion)
        \item Repeat until there are no more unfiltered windows
    \end{itemize}
\end{itemize}

The lists of seqlets across all tasks are then unionized. If a pair of seqlets overlaps by more than \texttt{overlap\_portion} (which defaults to 50\%), the seqlet with the higher score for its respective task is retained; as seqlet flanks are expanded in later steps of the algorithm, there is not a great concern of losing information from discarding overlapping seqlets.

\subsection{Metaclustering}
\label{sec:metaclustering}

After a unified list of seqlets is obtained across all tasks, seqlets are clustered into metaclusters according to their contribution scores across tasks. The contribution of a seqlet to a particular task is taken to be the total per-position contribution score for that task in the central \texttt{sliding\_window\_size} basepairs of the seqlet. However, before we can cluster these scores across tasks, it is necessary to transform the scores such that scores from different tasks are comparable (by this, we mean that the scores for different tasks can be on different scales depending on the confidence of the model in that particular task, so some kind of normalization is needed). We do this transformation by replacing each score by its percentile relative to the distribution of all scores. By default, the percentile is taken w.r.t. the distribution of the absolute values of scores, and a negative sign is applied if the original score was negative - thus, an extreme positive score would be transformed to +1.0, and an extreme negative score would be transformed to -1.0. If the user specified the \texttt{separate\_pos\_neg\_thresholds} option to be True, then the percentiles for positive values and negative values are computed separately (as opposed to taking the percentiles w.r.t. the absolute values).

We now describe the metaclustering algorithm in detail. We introduce the concept of an \emph{activity pattern}, which is a vector where each element is a 1, 0 or a -1. An activity pattern acts as a coarse-grained summary of a seqlet's activity across all the tasks. If there are $n$ tasks, then there are $3^n$ possible activity patterns. The goal of our metaclustering algorithm is to assign each seqlet to a distinct activity pattern, which will represent the metacluster.

Let $v'$ denote the vector containing the transformed scores for all tasks for some seqlet. We say that $v'$ is \emph{strongly compatible} with an activity pattern $p$ if, for every corresponding element $v'_i$ and $p_i$, either $v'_ip_i > \texttt{strong\_threshold}$ or $p_i = 0$ (that is, either $|v'_i|$ should exceed the \texttt{strong\_threshold} in the direction indicated by $p_i$, or $p_i$ should be 0; 0 is best understood as ``no constraint'' on a task). Similarly, we say that $v'$ is \emph{weakly compatible} with an activity pattern $p$ if, for every corresponding element $v'_i$ and $p_i$, either $v'_ip_i > \texttt{weak\_threshold}$ or $p_i = 0$. Note that the all-zeros activity pattern is compatible with all seqlets by both strong and weak thresholds.

How are \texttt{strong\_threshold} and \texttt{weak\_threshold} set? \texttt{strong\_threshold} is taken to be the most lenient CDF cutoff used across all tasks during per-task seqlet identification; the goal is to ensure that every seqlet has a transformed score of absolute value greater than \texttt{strong\_threshold} for at least one task. The \texttt{weak\_threshold} is defined by the user, but is set to be equal to \texttt{strong\_threshold} if the user specifies a \texttt{weak\_threshold} that exceeds \texttt{strong\_threshold}.

The metaclustering algorithm then proceeds as follows:
\begin{itemize}
    \item For each possible activity pattern, compute the number of ``strongly compatible'' seqlets. Note that a single seqlet is compatible with multiple activity patterns as some activity patterns are 0 for particular tasks.
    \item Filter out activity patterns that have fewer than \texttt{min\_metacluster\_size} compatible seqlets, or fewer than \texttt{min\_metacluster\_size\_frac} $\times$ \texttt{<total number of seqlets>} compatible seqlets (whichever is smaller). The default value for \texttt{min\_metacluster\_size} is 100, and the default value for \texttt{min\_metacluster\_size\_frac} is 0.01.
    \item Map each seqlet to the most specific activity pattern (that is, the activity pattern with the fewest zeros) with which it is weakly compatible. If a seqlet is compatible with two activity patterns of equal specificity, it is assigned to the one for which it has the higher absolute transformed score.
\end{itemize}

Emprically, we found that this metaclustering strategy produced metaclusters with relatively clean and distinct patterns relative to strategies like k-means clustering or community detection. In future versions of TF-MoDISco, we anticipate allowing the user to provide their own metaclusters that can supersede or act in combination with this metaclustering strategy. These metcalusters could be based on numerous alternative features such as the activations of the last hidden layer, the output label vector, or orthogonal signals such as methyation levels.

To improve the runtime of the algorithm, the number of seqlets in a metacluster is capped by taking a subset equal to the parameter \texttt{max\_seqlets\_per\_metacluster}.

\section{TF-MoDISco, Phase 2: Clustering Within a Metacluster}

In Phase 2 of TF-MoDISco, we perform subclustering of the seqlets in a metacluster using the importance scores for the subset of tasks that are relevant to the metacluster. We consider a task to be relevant to a metacluster if the activity pattern of the metacluster is nonzero for that task.

\subsection{Affinity Matrix Computation}

The fist step of Phase 2 is to compute affinities between every pair of seqlets, so that these affinities can be later clustered. Because this step scales with $O(N^2)$, where $N$ is the number of seqlets, we perform the computation in two steps: first, we compute a coarse-grained affinity matrix using a relatively quick method, and then for each seqlet, we compute affinities for \texttt{nearest\_neighbors\_to\_compute} nearest neighbors (as determined by the coarse-grained affinity matrix) using a more sophisticated but also more time-consuming method. The default value of \texttt{nearest\_neighbors\_to\_compute} is 500. \\

\subsubsection{\textbf{Coarse-Grained Affinity Matrix Computation: Gapped $k$-mer embedding}}

For the coarse-grained affinity matrix, a gapped $k$-mer embedding is derived for each seqlet, and the cosine similarity between seqlet embeddings is used as the affinity. The maximum similarity is taken over both the forward and reverse-complement versions of the seqlet, unless the user sets \texttt{revcomp} to False when calling TF-MoDISco. Note that this aspect of the workflow is highly specific to DNA sequence, and would have to be replaced with another way of computing affinity matrices if the user wishes to cluster patterns that are not DNA sequence (we may add support for this in future versions of TF-MoDISco). The gapped $k$-mer embedding proceeds as follows:

\begin{itemize}
\item For each seqlet, a single hypothetical contributions track is created by taking the sum of the hypothetical contributions for the tasks that are relevant for the metacluster's activity pattern (a task is ``relevant'' if it is nonzero in the activity pattern). If a task is negative in the activity pattern, the hypothetical contributions are multiplied by -1 before being added to the sum.
\item Gapped $k$-mers (potentially allowing for mismatches) are identified in the seqlet by scanning the one-hot encoded sequence.
\item For each gapped $k$-mer match, the ``score" of the match is taken to be the dot product of the one-hot encoded representation of the gapped $k$-mer and the portion of the summed hypothetical contributions track that it overlaps.
\item A total score for each gapped kmer is computed by summing the scores over all matches for that gapped kmer in the seqlet. This vector of total scores serves as the embedding.
\end{itemize}

For efficiency, the gapped kmer scoring is implemented using convolutional operation on the GPU.

\subsubsection{\textbf{Fine-grained Affinity Matrix Computation: The Continuous Jaccard Similarity}}
\label{sec:finegrained}

\begin{figure}[!h]
  \begin{center}
  \includegraphics[width=\linewidth]{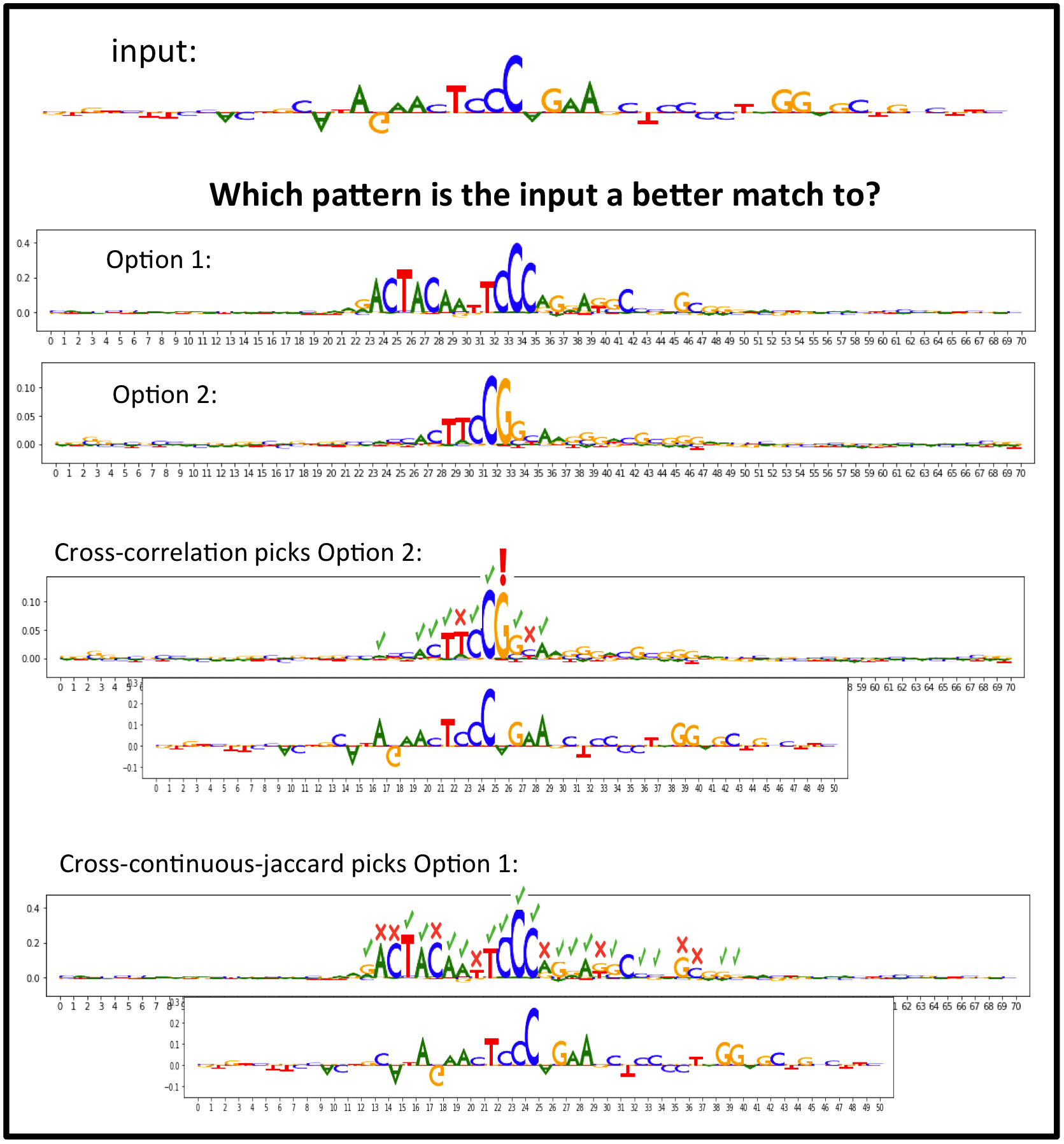}
  \caption[Continuous Jaccard similarity is preferable to cross-correlation for matching seqlets]{\textbf{Continuous Jaccard similarity is preferable to cross-correlation for matching seqlets}. Green checkmarks indicate matching positions.}
  \label{fig:crosscontinjaccard}
  \end{center}
\end{figure}

For the \texttt{nearest\_neighbors\_to\_compute} nearest neighbors of each seqlet as computed by the coarse-grained affinity matrix, we compute affinities using a more sophisticated (but computationally slower) similarity metric. Initially, we cross-correlated the importance score tracks of two seqlets and used the correlation at the best alignment as the similarity. However, using cross-correlation in this context has a drawback, which we explain here.

Consider the following toy example: we have a vector $v_1 = (-1,-1,-2,4,-1,-1,-1)$ of scores, and we are comparing it to two other vectors: $v_2 = (0, 0, 0, 4, 0, 0, 0)$ and $v_3 = (-1, -1, -2, 0, -1, -1, -1)$. The pearson correlation of $v_1$ with $v_2$ is equal to the cosine similarity between $v_1$ and $v_2$ after mean-normalization, and is 0.98. By contrast, the cosine similarity between $v_1$ and $v_3$ is 0.87.

To appreciate why the correlation between $v_1$ and $v_2$ is higher than the correlation between $v_1$ and $v_3$ even though $v_1$ and $v_2$ agree on the value of only one element while $v_1$ and $v_3$ agree on the value of 6, note that the formula for the correlation involves taking the elementwise product of terms. Thus, the formula has a polynomial degree of two, meaning that values of larger magnitude will have a disproportionately large influence on the correlation compared to values of smaller magnitude. In the context of comparing seqlets, this can cause issues as the importance scores at individual seqlets can be quite noisy, and a metric that is highly sensitive to the magnitudes of scores may be unreliable. Also note that the formula for correlation involves mean-normalization, which can again cause issues when outliers have a disproportionate impact on the mean.

As an alternative to correlation, we propose the following similarity measure, inspired by the Jaccard similarity. First, we define a notion of `intersection' and `union' for real numbers as follows:

\begin{equation}
\begin{split}
x \cap y &=  \min(|x|, |y|) \times \text{sign}(x) \times \text{sign}(y)\\
x \cup y &=  \max(|x|, |y|)
\end{split}
\end{equation}

Note that $x \cap y$ will have a positive sign if the signs of $x$ and $y$ agree, and a negative sign otherwise. We define the continuous Jaccard similarity between $v_1$ and $v_2$ as:

\begin{equation}
\label{eqn:continjaccard}
\begin{split}
\text{ContinuousJaccard}(v_1, v_2) = \frac{\sum_i v_1^i \cap v_2^i}{\sum_i v_1^i \cup v_2^i}
\end{split}
\end{equation}

Here, $v_1^i$ denotes the $i$th element of $v_1$. The Continuous Jaccard similarity will have a value of -1 if $v_1 = -v_2$, and a value of 1 if $v_1 = v_2$. Returning to our toy example, we find that $(-1,-1,-2,4,-1,-1,-1)$ has a Continuous Jaccard similarity of $\frac{4}{11}$ with $(0, 0, 0, 4, 0, 0, 0)$ and $\frac{7}{11}$ with $(-1, -1, -2, 0, -1, -1, -1)$.

As an illustration of the effectiveness of our similarity metric in practice, consider the example in Fig. ~\ref{fig:crosscontinjaccard}. The Continuous Jaccard similarity matches the input scores to the pattern that agrees with it at more positions compared to the match produced by using correlation.

What do we use as the input to our continuous jaccard similarity when we compare two seqlets? We use a combination of the hypothetical importances and the actual importances for all tasks deemed to be relevant for the seqlets. Specifically, let $H_{s_1}^t$ denote the hypothetical importance scores for seqlet $s_1$ and task $t$, and let $I_{s_1}^t$ denote the actual importance scores for seqlet $s_1$ and task $t$. Both $H_{s_1}^t$ and $I_{s_1}^t$ have dimensions $l_{s} \times 4$, where $l_{s_1}$ is the length of a seqlet. Let $\hat{H}_{s_1}^t$ and $\hat{I}_{s_1}'^t$ denote L1-normalized version of $H_{s_1}^t$ and $I_{s_1}^t$ (in other words, the sum of the absolute values of the elements in each of $\hat{H}_{s_1}^t$ or $\hat{I}_{s_1}'^t$ totals 1). We form a new matrix $S_1$ by concatenating the matrices $\hat{H}_{s_1}^t$ and $\hat{I}_{s_1}'^t$ for all relevant tasks along the second dimension. If $m$ is the total number of relevant tasks, the final dimensions of $S_1$ will be $l_{s} \times 4\times 2 \times m$ (the factor of 2 is introduced because we are using both actual importances and hypothetical importances for every task).

The matrices $S_1$ and $S_2$ are then padded with zeros and compared at all alignments where the non-padded matrices overlap at at least $\texttt{min\_overlap\_while\_sliding} l_s$ positions using the Continuous Jaccard similarity metric (the default value of \texttt{min\_overlap\_while\_sliding} is $0.7$). The maximum Continuous Jaccard similarity over all possible alignments is take to be the similarity between $s_1$ and $s_2$. This maximum is computed over both the forward and reverse-complement versions of the seqlet, unless the user sets \texttt{revcomp} to False when calling TF-MoDISco.\\

\subsubsection{\textbf{Noise filtering}}

\begin{figure}[!h]
  \begin{center}
  \includegraphics[width=\linewidth]{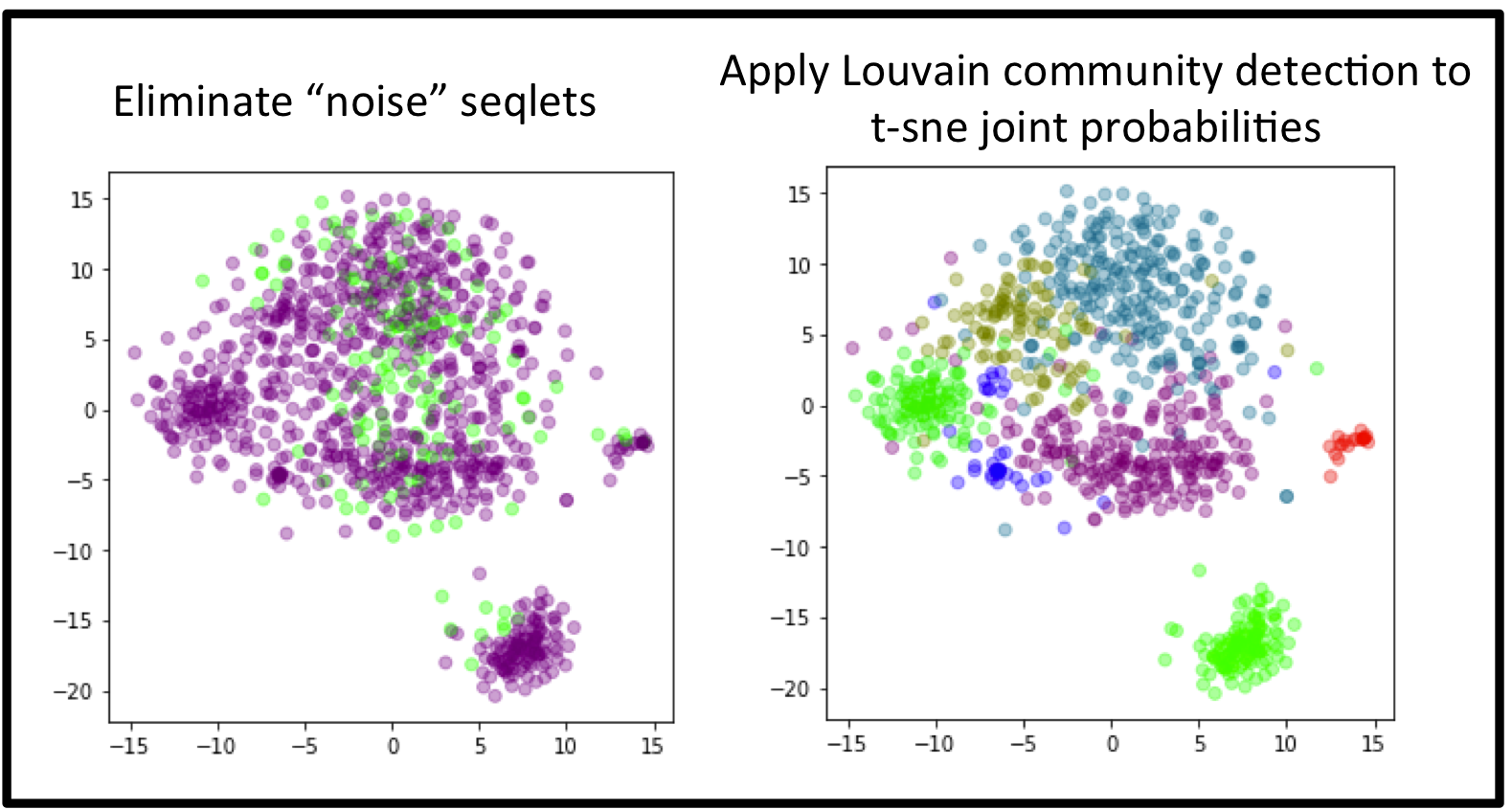}
  \caption[Noise filtering and seqlet clustering]{\textbf{Noise filtering and seqlet clustering.} Left: Seqlets with poor spearman correlation between the coarse-grained and fine-grained affinities were discarded. Right: The remain seqlets were clustered to form an initial set of motifs.}
  \label{fig:seqletclustering}
  \end{center}
\end{figure}

We found that seqlets which had very poor correlation between the coarse-grained affinities and the fine-grained affinities tended to resemble noise. We leverage this observation and discard all seqlets for which the spearman correlation between coarse-grained and fine-grained affinities is below \texttt{affmat\_correlation\_threshold}, which defaults to 0.15. See the left panel of Fig. ~\ref{fig:seqletclustering} for a tsne-embedding showing the seqlets eliminated by our noise filtering step.

\subsection{Clustering the Affinity Matrix}
\label{sec:clusteringtheaffmat}

Now that we have an affinity matrix, the next step is to obtain clusters. We first perform a transformation of the affinity matrix intended to adapt the notion of distance to the local density of the data.

\subsubsection{\textbf{Adapting the Notion of Distance to the Local Density of the Data}}
A challenge with using the original affinity matrix is that the notion of what is `close' can vary with the motif cluster. For example, a weak motif might have more weak matches present in the data compared to the strongest motif, so what we might consider a good similarity score for the weak motif may not be appropriate for the strong motif. One way to address this is to adapt the notion of `close' to the local density of the data; thus, if a seqlet has a lot of neighbors with high similarity, we can afford to be more stringent in our definition of `close'. By contrast, if a seqlet's nearest neighbors have relatively low similarity, we may have to be more lenient.

To achieve this, we borrow from t-sne \cite{maaten2008visualizing}. The first step of t-sne is to compute a matrix of conditional probabilities $p_{j|i}$ that represent the likelihood that seqlet $s_j$ is picked conditioned on $s_i$ having already been picked, where $p_{j|i}$ is proportional to a Gaussian centered on $i$:

\begin{equation}
\label{eqn:tsnecondprob}
p_{j|i} \propto \exp(-\beta_i d_{s_i s_j})
\end{equation}

Where $d_{s_i s_j}$ is the distance between seqlet $s_i$ and $s_j$. Crucially, the value of $\beta_i$ is tuned such that the $p_{j|i}$ distribution achieves a target perplexity. A distribution in which the $k$ nearest seqlets of $i$ each had probability $\frac{1}{k}$ and every other seqlet had probability zero would have a perplexity of exactly $k$. Thus, perplexity can roughly be thought of as the size of the neighborhood of $i$ in which $p_{j|i}$ is high. In our experiments, we used a perplexity of 10 as this proved effective at pulling out small clusters.

One caveat is that the aforementioned transformation uses distances $d_{s_i s_j}$, but the Continuous Jaccard metric outputs a similarity. We map our similarities to distances using the formula:

\begin{equation}
\label{eqn:afftodist}
y = \log\left(\frac{1}{0.5\max(x,0)} - 1\right)
\end{equation}

\begin{figure}[!h]
  \begin{center}
  \includegraphics[width=0.6\linewidth]{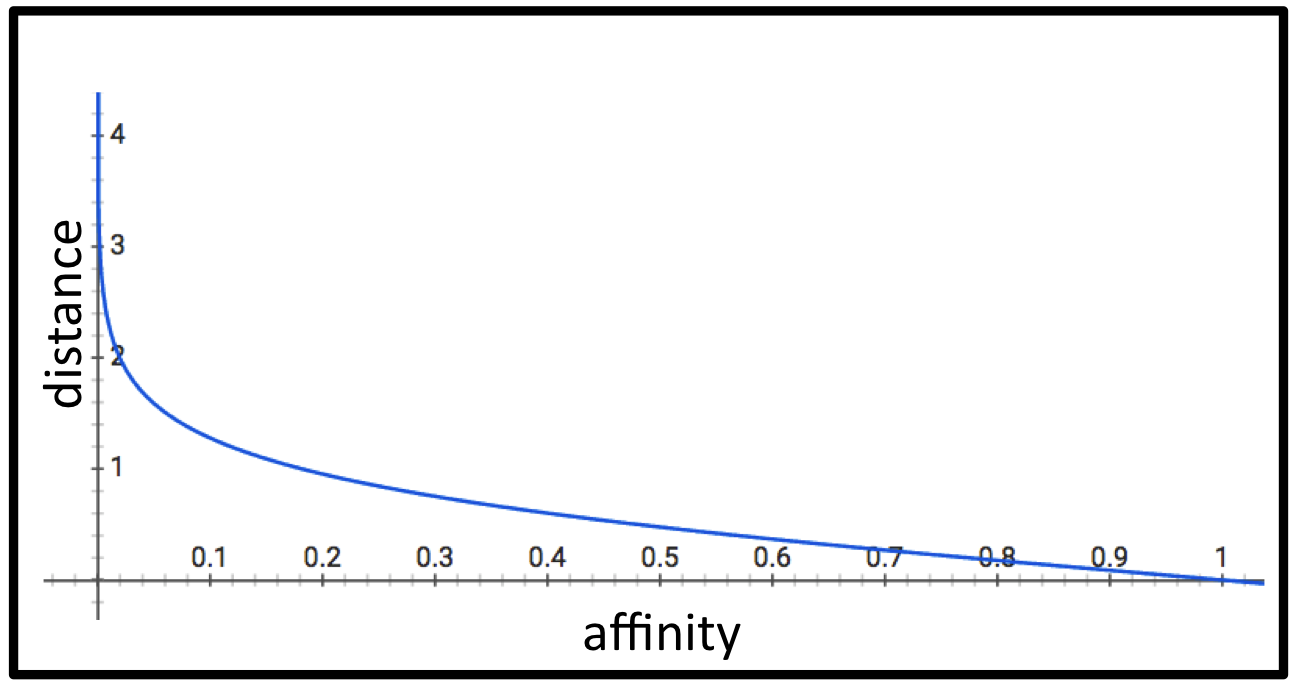}
  \caption[Plot of $y = \log\left(\frac{1}{0.5\max(x,0)} - 1\right)$, used to map affinities to distances]{\textbf{Plot of $y = \log\left(\frac{1}{0.5\max(x,0)} - 1\right)$, used to map affinities to distances.}}
  \label{fig:afftodistfunc}
  \end{center}
\end{figure}

See Fig ~\ref{fig:afftodistfunc} for an illustration of the function. We found that this transformation worked better than naively using $y = 1-x$, because when the similarity approaches 0, the distance tends to infinity (in practice, the lowest similarity between two seqlets tends to be around 0 rather than around -1).

Note that at no point do we compute a lower-dimensional t-sne embedding for clustering purposes. We work directly with the similarities computed in the higher-dimensional space and convert them into a conditional probability matrix. We then symmetrize the conditional probability matrix by computing the joint probabilities $p_{ij}$ as follows:

\begin{equation}
p_{ij} = \frac{p_{j|i} + p_{i|j}}{2N}
\end{equation}

Where $N$ is the total number of seqlets. This formula for the joint probability of picking both $s_i$ and $s_j$ assumes that the first seqlet $s_1$ is picked at random and the second seqlet $s_2$ is picked according to $p_{s_2 | s_1}$.\\

\subsubsection{Clustering using Louvain or Leiden Community Detection}

After the affinity matrix has been transformed to adapt to the local density of the data, we peform community detection. TF-MoDISco supports both Louvain and Leiden community detection; the choice of which algorithm to use is controlled by the \texttt{use\_louvain} argument of the \texttt{TfModiscoSeqletsToPatternsFactory} class). We describe the workflow associated with each community detection algorithm below:

\textbf{Clustering using Louvain}

A popular choice for community detection is the Louvain algorithm \cite{blondel2011the}. Louvain community detection scales well, does not require the user to pre-specify a number of clusters, and has been used successfully in other biological applications such as PhenoGraph \cite{levine2015data}. However, Louvain community detection is not deterministic. We found that if Louvain community detection was applied directly to the joint probabilities described above, the results were somewhat unstable. To reduce this this instability, we perform consensus clustering, in that we run Louvain community detection \texttt{louvain\_membership\_average\_n\_runs} times (default of 200 times) and define the new affinity between two seqlets to as the fraction of times Louvain community detection placed the two seqlets in the same cluster at the highest level of the hierarchy.

Once we have the new affinity matrix derived from consensus clustering, we perform Louvain community detection again and use the communities produced at the lowest level of the hierarchy as our clusters. To add an additional layer of robustness to random seed, we try different random seeds until we see no improvement in modularity over 50 consecutive runs of Louvain, and then take the clustering that produced the best modularity. We used an implementation of Louvain community detection that is built on the one present in the PhenoGraph package \cite{levine2015data}. See the right panel of Fig. ~\ref{fig:seqletclustering} for an example of a tsne-embedding visualizing the final clusters picked out by Louvain community detection. 

\textbf{(New in version 0.5.6) Clustering using Leiden}

Despite its popularity, \citet{leiden} found that the Louvain algorithm is prone to producing arbitrarily badly-connected communities. As an alternative, they proposed the Leiden community detection algorithm. As of version 0.5.6, TF-MoDISco supports using Leiden community detection instead of Louvain. Empirically, we observed that the consensus clustering approach described for Louvain did not work well with Leiden; thus, our pipeline for Leiden community detection does not include this. Instead, we perform Leiden community detection with 50 different random seeds and take the clustering that produces the best modularity. For each random seed, the Leiden algorithm is iterated until there is no improvement.

\subsection{Seqlet Aggregation}

We aggregate the seqlets within each cluster to generate `motifs'. To do this, we sort seqlets in descending order of the total magnitude of their importance on all the relevant tasks and merge them together successively in a greedy fashion. Specifically, when merging a seqlet into the aggregated motif, we align the seqlet to all positions where at least \texttt{min\_overlap\_while\_sliding} of the seqlet overlaps the motif, and pick the alignment that produces the best Continuous Jaccard similarity. The aggregated motif is then updated by averaging the values of each data track at each position over all the seqlets aligning to that position. This is done until there are no more seqlets left for merging.

\subsection{Motif boundary editing}
\label{sec:motifboundarediting}

The boundaries of the motifs are then editing in three steps, as described below and illustrated in step (iv) of \textbf{Fig. ~\ref{fig:methodsumary}}.

\subsubsection{\textbf{(a) Trim away flanks with weak support}}

Once all the seqlets have been collapsed into a single motif, the boundaries of the motif are trimmed to retain only those positions that have a good number of seqlets covering them. Specifically, at each position in the motif we count the number of seqlets for which the center of the seqlet is aligned to that position. We calculate the number of seqlet centers at the position with the most seqlet centers, and define our threshold to be \texttt{frac\_support\_to\_trim\_to} (default 0.2) of this value. If this threshold is below \texttt{min\_num\_to\_trim\_to} (default 30), we set the threshold to \texttt{min\_num\_to\_trim\_to} instead. Starting from the ends of the motif, we discard seqlets whose center aligns to positions where the total number of seqlet centers does not not meet the threshold. We stop trimming from a particular end when we encounter a position that passes the threshold.

\subsubsection{\textbf{(b) Expand seqlets to fill their motifs}}

Once the motif has been trimmed down to retain only those seqlets aligning to positions with good coverage, the ends of the remaining seqlets are expanded on either side to fill the boundaries of the motif as illustrated in step (iv.b) of \textbf{Fig. ~\ref{fig:methodsumary}}. Note that this expansion step improves the quality of the aggregated motif, because each position in the motif now has information from all the seqlets aligned to the motif.

\subsubsection{\textbf{(c) Center motifs and standardize lengths}}

For subsequent steps, it is desirable to have motifs that are all of uniform size, even if it means including some uninformative positions in the motif (these uninformative positions can be trimmed away later). To make the motifs have uniform size, each motif is then trimmed to the \texttt{trim\_to\_window\_size} bp window (default 30) that has the highest total magnitude of both real and hypothetical importance for all the relevant tasks, and the seqlets in the \texttt{trim\_to\_window\_size} bp window are then expanded by \texttt{initial\_flank\_to\_add} bp on either size, producing motifs of length $\texttt{trim\_to\_window\_size} + 2\times\texttt{initial\_flank\_to\_add} $bp.

\subsection{Discard motifs that disagree with metacluster activity pattern}

After motif boundary editing, the aggregated motif's contribution score pattern may disagree with that of the activity pattern of the metacluster. We have observed that this can be because the original seqlet center was at the noisy flank of a motif where the importance scores on the flank had a different sign than the importance score of on the motif. We filter these motifs out.

\subsection{Repeat Seqlet Clustering}

In the previous step, the seqlet boundaries are edited and expanded. Often, this can result in each seqlet incorporating additional informative positions that could improve the clustering. Thus, we take the seqlets from all the processed motifs and subject them to a second round of clustering as illustrated in Fig. ~\ref{fig:methodsumary}. We found that in practice, this second round of clustering improved the quality and stability of results. In principle, more rounds can be performed, but the current default is to do only one additional round.

\section{TF-MoDISco Phase 3: Post-processing of clusters}

\subsection{Identifying spurious merging}

We do a final check to make sure that our Louvain community detection process did not merge two motifs together that one could argue belong in different clusters. To do this, we alter the implementation of Louvain community detection such that, during initialization, each point is randomly assigned to one of two clusters (by contrast, the original initialization assigns each point to its own clusters); this guarantees that at the end of community detection, at most two clusters are returned. We then run Louvain with different random seeds until no improvement in modularity is observed over 20 consecutive random seeds; the clustering that had the best modularity is returned. We refer to this process as ``Louvain diclustering".

The spurious merging detection proceeds recursively as follows:

\begin{itemize}
    \item For each motif, we first check to see whether the motif contains more than \texttt{final\_min\_cluster\_size} (default 30) seqlets; if it does, we inspect the motif for spurious merging. we look at the constituent seqlets and compute the similarity between seqlets using the Continuous Jaccard Similarity metric (\textbf{Eqn. ~\ref{eqn:continjaccard}}). This is similar to the process used to compute the original fine-grained affinity matrix, except that we look at all seqlet-seqlet pairs and we do not need to look at different alignments of the seqlets, as we rely on the seqlet alignment that was used to aggregate the seqlet into the motif.
    \item Given the seqlet-seqlet similarity matrix for seqlets in a motif, we run Louvain diclustering. If the diclustering produces more than one subcluster, we aggregate the seqlets within the subclusters to produce sub-motif.
    \item We check whether the sub-motifs are substantially dissimilar by looking at their Pearson correlation. If the Pearson correlation is less than \texttt{threshold\_for\_spurious\_merge\_detection} (default 0.8), we declare the two sub-motifs to be dissimilar and repeat the spurious merging detection on each sub-motif. Otherwise, we terminate the recursion.
\end{itemize}

\subsection{Merging Redundant patterns}

Having performed spurious merging detection, we now collapse similar motif clusters together to obtain a final list of non-redundant motifs. We will use two points of information when deciding whether to merge: one is the maximum cross-correlation (after mean and magnitude normalizing the score tracks) over all alignments of the motifs that overlap by at least \texttt{min\_overlap\_while\_sliding} (default 0.7), and the other is a measure of similarity that takes into account how tightly packed the motif is. We find that cross-correlation performs better than cross Continuous Jaccard at this stage because the motifs are aggregated over several seqlets and thus the scores are less noisy. We describe the second measure in more detail in the next section.\\

\subsubsection{\textbf{Density-sensitive similarity}}

In Sec. ~\ref{sec:clusteringtheaffmat}, we used conditional probabilities inspired by t-sne to adapt our notion of distance to the local density of the data. We wish to leverage a similar idea when computing the similarity between two motif clusters. Let $m_i$ denote a motif. For all seqlets $s_j$ that were clustered to form the list of motifs, we define $p_{j|i}$ as:

\begin{equation}
p_{j|i} \propto \exp(-\beta_i d_{s_j m_i})
\end{equation}

Note that this formula is analogous to eqn. ~\ref{eqn:tsnecondprob}. Here, $d_{s_j m_i}$ is the distance from seqlet $s_j$ to motif $m_i$, computed using the Continuous Jaccard similarity subject to the transformation in eqn. ~\ref{eqn:afftodist}. The purpose of this is to calculate the value of $\beta_i$ such that the perplexity of $p_{s|m_i}$ is equal to the number of seqlets that aligned to motif $m$. We then define the density-sensitive similarity as:

\begin{equation}
\text{DensitySensitiveSim}_{m_i m_j} = \exp(-\max(\beta_i, \beta_j) d_{m_j m_i})
\end{equation}

As high values of $\beta$ indicate tighter packing, this can be thought of using the more tightly-packed motif cluster to calculate the density-adjusted similarity to the other motif.\\

\subsubsection{\textbf{Iterative motif merging}}

Once we have our motif similarities from cross-correlation as well as our density-sensitive similarities, we are ready to merge motifs together. We wish to collapse highly similar motifs together, but our idea of similarity may not be transitive (that is, if $m_1$ is similar to $m_2$ and $m_2$ is similar to $m_3$, it is not necessary that $m_1$ is sufficiently similar to $m_3$ that we would be comfortable merging $m_1$, $m_2$ and $m_3$ together). To avoid merging dissimilar motifs together, we will perform our motif merging use a combination of two criteria: one criterion for whether two motifs are similar enough to attempt to merge them, and another for whether two motifs are sufficiently dissimilar that we do not want them merged together.

We use an iterative algorithm as follows: Let $S_i$ denote the set of motifs that are to be merged with motif $i$, including $i$ itself; at the beginning of an iteration, this is just a set containing the single element $i$. We enumerate all possible motif pairs $(i,j)$ and iterate over the pairs in descending order of their cross-correlation similarity. If the combination of the cross-correlation similarity and the density-sensitive similarity meet our predefined criterion potentially merging $i$ and $j$, we proceed to iterate over all the motifs currently in $S_i$ and $S_j$ and make sure that no pair of motifs meet our criterion for incompatibility. If no incompatibilities are found, we update our sets to reflect the fact that all the motifs in $S_i$ and $S_j$ are going to be merged together. At the end of the iteration, we aggregate all the seqlets in the motifs of each set and repeat the process until we reach an iteration where no motifs are merged together.

The criteria for motifs being considered similar or dissimilar can be modified by the user. Let $p$ represent the density-sensitive similarity between two motifs, and let $c$ represent the average cross-correlation-based similarity. The default criterion for motifs to be considered similar is that at least one of the following must be true: ($p > 0.0001$ and $c > 0.84$), or ($p > 0.00001$ and $c > 0.87$), or ($p > 0.000001$ and $c > 0.90$). The default criterion for motifs to be considered dissimilar is that at least one of the following must be true: ($p < 0.1$ and $c < 0.75$), or ($p < 0.01$ and $c > 0.8$), or ($p < 0.001$ and $c < 0.83$), or ($p < 0.0000001$ and $c < 0.9$). These defaults were set based on empirical observations. Future versions of TF-MoDISco will likely refine this motif merging step.

\subsection{Reassign seqlets from small clusters to the dominant clusters}

We disband motifs for which the number of seqlets in the motif falls below \texttt{final\_min\_cluster\_size} (default 30). The seqlets in the disbanded motifs are assigned to whichever of the remaining motifs they have the highest affinity to (as measured by Continuous Jaccard similarity), provided that the similarity exceeds \texttt{min\_similarity\_for\_seqlet\_assignment} (default 0.2).

\subsection{Final flank expansion}

The motifs are ultimately expanded to reveal potential flanking patterns. In this work, we expand the motifs to include an additional \texttt{final\_flank\_to\_add} (default 10) bp on either side, for a total of $\texttt{trim\_to\_window\_size} + 2\times(\texttt{initial\_flank\_to\_add} + \texttt{final\_flank\_to\_add})$bp. Note that flanks can be trimmed as desired if they are deemed to be uninformative.

\section{(New in v0.5.6) Leveraging MEME to initialize the TF-MoDISco clusters}

As of v0.5.6, it is possible to initialize the TF-MoDISco clusters using MEME. The initialization proceeds as follows: the sequences corresponding to the seqlets in a metacluster (identified as described in \textbf{Sec. \ref{sec:seqletidentification} \& \ref{sec:metaclustering}}) are written to a fasta file. MEME is run on this fasta file, and the resulting PWMs are used to scan the seqlets. For each PWM and seqlet, the maximum log-odds score across all possible alignments is computed. If this maximum log-odds score exceeds the Bayes optimal threshold specified by MEME for the PWM, the seqlet is considered to have a match to the PWM. Seqlets that contain no matches to \emph{any} PWM are assigned to their own cluster. The remaining seqlets are each assigned to a cluster corresponding to the PWM for which they had the strongest match by log-odds score.

This MEME-based clustering is then leveraged in the TF-MoDISco workflow in two places. First, the fine-grained similarity (\textbf{Sec. \ref{sec:finegrained}}) is computed not just on the set of nearest-neighbors that have the highest coarse-grained similarity across all seqlets, but also on the set of nearest-neighbors that have the highest coarse-grained-similarity within each initialized cluster. Second, the MEME-based clustering is used to initialize Leiden community detection (\textbf{Sec. \ref{sec:clusteringtheaffmat}}; this workflow is not supported when Louvain community detection is used instead of Leiden).

In principle, any strategy for identifying motifs could be applied to the seqlets and used to create an initial clustering. We note that applying motif discovery to the sequences underlying the seqlets - as opposed to applying motif discovery to the original full sequences - can often greatly improve the sensitivity of the motif discovery by focusing attention on regions that are highly likely to contain motifs.

An example of how to use the workflow involving MEME-based initialization is demonstrated in \href{https://github.com/kundajelab/tfmodisco/blob/ffdc70bf08852a18f41ba684a58f54cbd5f8c1d0/examples/H1ESC_Nanog_gkmsvm/TF\%20MoDISco\%20Nanog.ipynb}{\color{blue}this jupyter notebook}.

\subsection*{Author Contributions}

Avanti Shrikumar conceived of and was the primarily developer of TF-MoDISco. Katherine Tian, Anna Shcherbina, \v{Z}iga Avsec and Surag Nair contributed features to the TF-MoDISco codebase. Anna Shcherbina proposed the idea of fast affinity matrix computation using kmers. Abhimanyu Banerjee, Mahfuza Sharmin and \v{Z}iga Avsec applied versions of TF-MoDISco immediately leading up to v0.5.6.5 on biological data and identified areas of improvement. Anshul Kundaje provided guidance and feedback. Avanti Shrikumar wrote this technical note.

\subsection*{Acknowledgments}

We thank Stefan Holderback, Han Yuan and Charles McAnany for their feature suggestions and pull requests to the TF-MoDISco codebase. We thank Nasa Sinnott-Armstrong for writing an initial version of GPU-based cross-correlation which was used in early prototypes of TF-MoDISco. We thank Chuan Sheng Foo for exploring k-means based clustering approaches for early prototypes of TF-MoDISco. We thank Chuan Sheng Foo, Johnny Israeli, Peyton Greenside and Irene Kaplow for evaluating early prototypes of TF-MoDISco on biological data.

\bibliographystyle{plainnat}
\bibliography{bibliography.bib}

\end{document}